\documentclass[final,5p,times,twocolumn,number]{elsarticle}

\usepackage{amssymb}
\usepackage{amsmath,amsfonts}
\usepackage{algorithmic}
\usepackage{algorithm}
\usepackage{graphicx}
\usepackage{textcomp}
\usepackage{xcolor}
\usepackage{booktabs}
\usepackage{multirow}
\usepackage{hyperref}
\usepackage{bm}
\usepackage{float}
\usepackage{stfloats}
\usepackage{pdfpages}
\usepackage{orcidlink}

\begin{document}

\begin{frontmatter}

\title{StaX-Net: Stackelberg Interleaved Asymmetric Network\\for Cross-Modality Medical Image Translation\tnoteref{t1}}

\author[first]{Amin Golnari \orcidlink{0000-0002-3841-0588}\corref{cor1}}
\ead{amin.g@vinuni.edu.vn}
\author[first,second,third]{Saeid Sanei \orcidlink{0000-0002-1446-5744}}
\ead{saeid.s@vinuni.edu.vn}
\cortext[cor1]{Corresponding author}

\affiliation[first]{organization={College of Engineering \& Computer Science, VinUniversity},
            city={Hanoi},
            country={Vietnam}}
\affiliation[second]{organization={College of Health Science, VinUniversity},
            city={Hanoi},
            country={Vietnam}}
\affiliation[third]{organization={Institute of Psychiatry, Psychology \& Neuroscience, King's College London},
            country={UK}}

\begin{abstract}
Cross-modality medical image translation is challenged by anatomically inconsistent synthesis that can compromise clinical diagnosis. We propose StaX-Net, a Stackelberg interleaved asymmetric network that incorporates a partially coupled encoder-discriminator structure, and a 2D feature aggregation and spatial transformation (FAST) layer that distills target-supervised priors under the learning using privileged information paradigm. The shared blocks between the generator and discriminator enhance architectural feature alignment. To stabilize the leader-follower training, a Stackelberg-inspired asymmetric optimization scheme employs explicit shared-parameter freezing and asymmetric learning rates. The FAST layer distills target-domain priors during training and applies a source-only transform at inference. This constrains the synthesis process through target-informed feature representations relative to standard generative architectures. The code is available on \href{https://github.com/amingolnari/Demo-StaX-Net}{GitHub} for public use.
\end{abstract}

\begin{keyword}
Coupled-GAN \sep Cross-Modality Synthesis \sep LUPI \sep Medical Image Translation \sep Stackelberg Games \sep Weight Sharing
\end{keyword}

\end{frontmatter}

\section{Introduction}
The translation of medical image modalities, specifically for synthesis of high-fidelity magnetic resonance imaging (MRI) representations from computed tomography (CT) inputs, has become a high impact research focus in computational radiology and clinical decision support. While CT provides excellent spatial contrast for hard tissues, it suffers from limited soft-tissue contrast compared to MRI, which excels at resolving subtle anatomical pathologies \cite{alamir2022role}. The clinical imperative driving cross-modality translation is profound: enabling the extraction of MRI-informed anatomical priors directly from CT acquisitions would reduce patient exposure to multiple imaging sessions, lower healthcare costs, and accelerate diagnostic workflows \cite{saad2024survey}.

Generative adversarial networks (GANs) have demonstrated remarkable capability in image synthesis and cross-domain translation \cite{chakraborty2023ten, zhao2021generative}. However, traditional GAN architectures frequently suffer from semantic misalignment, representation drift, and the generation of anatomically implausible hallucinations \cite{li2026som, cohen2018distribution, chakraborty2023ten}. In medical imaging, where an anatomical hallucination carries severe clinical risk and can directly compromise downstream tasks such as radiation therapy planning or fracture diagnosis, traditional adversarial instability remains a significant concern \cite{zhang2022general}.

Here, we present a framework that addresses these limitations through three integrated components designed to resolve gradient conflicts and feature misalignment inherent in standard coupled networks. First, unlike conventional cooperative weight-sharing models, we introduce an interleaved encoder-discriminator structure that tightly couples the adversarial networks through shared multi-scale computational blocks, while strictly maintaining separate pathway-specific activations and task-specific dedicated layers. 

Second, we replace simultaneous adversarial updates with a Stackelberg-inspired asymmetric optimization scheme. By enforcing strict asymmetric parameter ownership, the shared blocks are frozen during the generator update. This ensures the discriminator acts as a definitive leader in shaping the shared representation, preventing the generator from inducing conflicting gradient updates. Finally, during learning using privileged information (LUPI) paradigm, a 2D feature aggregation and spatial transformation (FAST) layer is integrated into the generator. Instead of serving as a direct input, the privileged reference (MRI) is utilized exclusively via a pathway-specific projection to guide the FAST layer's objective. This distills target-domain structural priors into a dynamic, source-only feature transformation, thereby acting as a structural regularizer that guides the synthesis process away from anatomically implausible representations.

\section{Related Work}

\subsection{Architectural Evolution: From Disjoint to Coupled GANs}
The evolution of GAN technology initially relies on disjoint parameter sets $\theta_G$ and $\theta_D$ \cite{goodfellow2014generative, arjovsky2017wasserstein, gulrajani2017improved, bermano2022state, chakraborty2023ten}, which often leads to computational inefficiency, divergent feature spaces, and mode collapse \cite{zhang2025mode, khayatkhoei2021spatial}. While recent works successfully mitigate mode collapse using auxiliary discriminators and Siamese networks \cite{boukhennoufa2023novel}, resolving feature divergence requires structural constraints. To this end, coupled architectures (e.g., CoGAN \cite{liu2016coupled}, CoCoGAN \cite{wang2019conditional, figueira2022survey}, and ALI/BiGAN \cite{donahue2016adversarial, dandi2020generalized}) leverage parameter sharing to enforce structural constraints and high-level semantics alignment. Furthermore, explicitly interleaving parameters between $G$ and $D$ within a single pipeline without inducing gradient conflict remains a largely uncharted domain.

\subsection{Game-Theoretic Training Dynamics}
Traditional GAN training relies on a simultaneous, zero-sum minimax game that frequently fails to reach a stable Nash equilibrium, producing non-convergent cyclical dynamics \cite{mescheder2018which, fedus2018many}. Stackelberg game formulations, widely successful in complex decision-making domains \cite{sinha2018stackelberg}, offer a principled framework for asymmetric optimization, where a dominant leader commits to a strategy and the follower responds \cite{zhang2018stackelberg, fiez2020implicit}. Extensions such as batch policy learning \cite{zhou2023stackelberg}, generalized Nash equilibrium problems (GNEP) \cite{akaligwo2023regularized}, and adversarial regularization via unrolled optimization \cite{zuo2021adversarial} have been proposed. Additionally, techniques like FreezeD \cite{mo2020freezed} demonstrate that selectively freezing discriminator layers stabilizes training, providing an empirical precedent for the selective freezing strategy employed herein.

\subsection{Weight Sharing and Feature Alignment}
In medical imaging, GANs have been applied to brain tumor segmentation \cite{ali2024brain}, MRI reconstruction \cite{ahmed2021fame}, and cross-modality synthesis \cite{chen2022beyond}. The FAME framework employed a globally shared encoder backbone in federated adversarial learning for MRI reconstruction \cite{ahmed2021fame}. IB-GAN integrated information bottleneck constraints into cycle-consistency GANs for domain adaptation in medical imaging \cite{chen2022beyond}.

CrossNet introduced latent cross-consistency constraints for unpaired image translation, sharing encoder/decoder weights to enforce consistent semantics \cite{sendik2019crossnet}. These works collectively demonstrate that parameter coupling can provide benefits for feature alignment in various applications such as MRI reconstruction, tumor segmentation, and cross-modality synthesis \cite{ali2024brain, ahmed2021fame, chen2022beyond, sendik2019crossnet}. However, parameter sharing alone does not specify how shared transformations are arranged across pathways or assigned to competing objectives. StaX-Net addresses this specific challenge through the stage-wise reuse of shared transformations combined with an explicit asymmetric update rule. The discriminator strictly controls the optimization of the shared parameters, while the generator adapts only its dedicated synthesis modules, distinguishing our framework from conventional coupled architectures.

\subsection{Stability, Lipschitz Continuity, and Convergence}
Stability in GANs is tied to the Lipschitz continuity of the discriminator \cite{arjovsky2017wasserstein}. WGAN-GP enforces this via gradient penalties \cite{gulrajani2017improved}, while spectral normalization provides an alternative mechanism \cite{liu2020lipschitz}. Shared backbones can act as implicit regularizers by constraining the feature space to a common representation, limiting divergence between $G$ and $D$ features \cite{liu2016coupled}. This restriction of the optimization landscape can contribute to training stability \cite{iglesias2022survey}.

\section{Proposed Methodology}

\begin{figure*}[htbp]
    \centering
    \includegraphics[width=0.97\textwidth, trim= 2cm 1.5cm 2cm 1.5cm, clip]{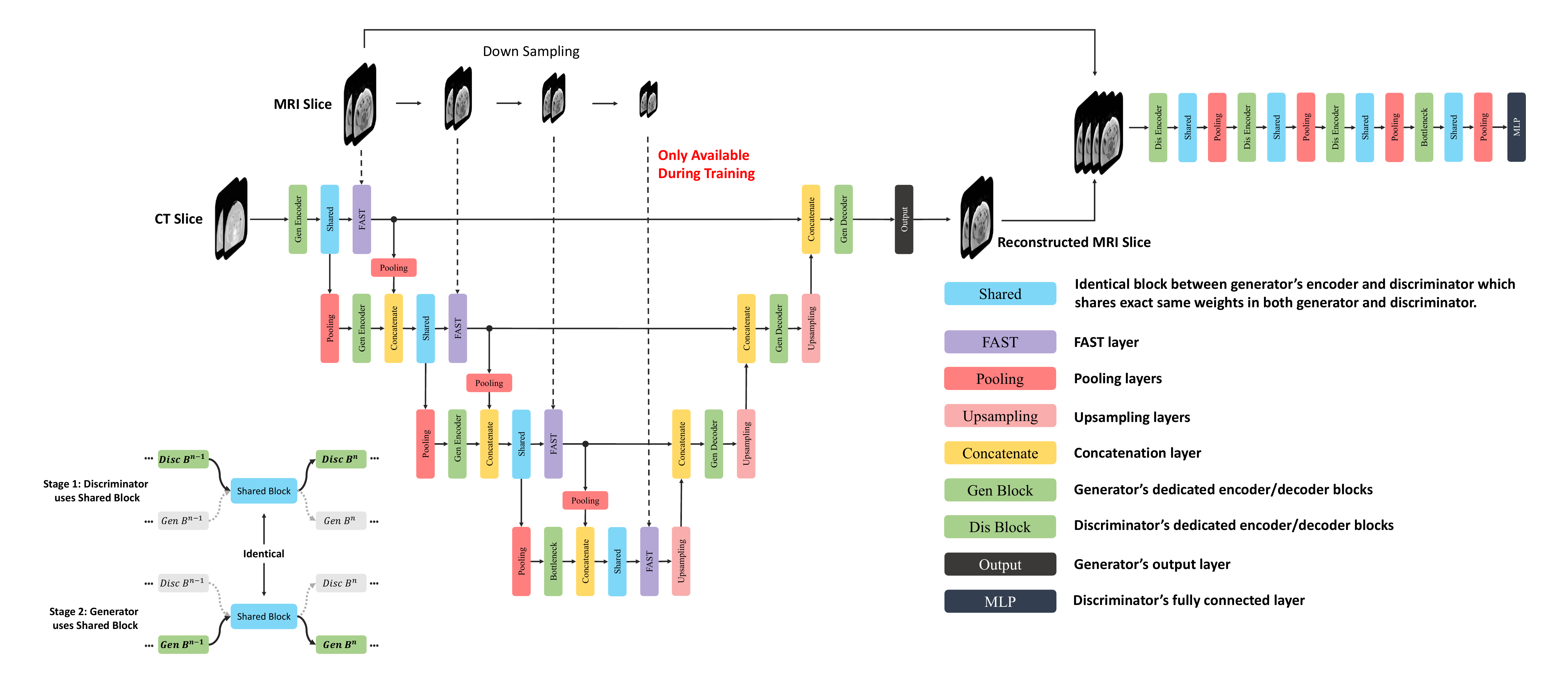}
    \caption{Overall architecture of the proposed StaX-Net framework. The generator (left) translates input CT slices into reconstructed MRI representations, while the discriminator (right) evaluates real versus generated MRI slices. A defining characteristic of this architecture is the tight coupling at each spatial resolution stage via identical shared blocks (blue). Structurally, the generator employs dual routing pathways at each scale, incorporating multi-scale MRI priors into the FAST layers (purple) exclusively during training, and utilizing skip connections that directly link FAST outputs to the decoding pathway. Under the proposed Stackelberg-inspired asymmetric optimization, the shared weights are explicitly updated by the discriminator during the first training phase (leader). In the second phase (follower), the generator utilizes these exact, newly updated shared weights as fixed (frozen) feature extractors, optimizing only its dedicated modules.}
    \label{fig:overall_architecture}
\end{figure*}

\subsection{Coupled Encoder-Discriminator Structure}
To address the limitations of disjoint parameter spaces, StaX-Net employs a multi-stage interleaved encoder structure. To explicitly distinguish our architecture from popular parameter-sharing designs, we embed $K$ blocks at corresponding stages of the generator and discriminator pathways. Let $\theta_{\text{shared}} = \{\theta_{\text{shared}}^{(k)}\}_{k=1}^{K}$ denote these shared blocks. At stage $k$, the generator and discriminator process their distinct incoming feature representations through the same parameterized block $\theta_{\text{shared}}^{(k)}$.

Importantly, sharing in StaX-Net does not imply that $G$ and $D$ process an identical feature tensor, nor does it reduce the architecture to a single shared encoder followed by independent branches. The resulting activations remain pathway-specific. Between consecutive shared stages, the networks apply dedicated transformations with distinct task-specific parameters: $\theta_{G_{\text{ded}}}$ for synthesis and $\theta_{D_{\text{ded}}}$ for classification.

During the discriminator-controlled update, gradients from the adversarial objective update the shared-block parameters. During the generator update, these shared blocks remain active in the forward pathway as fixed feature extractors. The gradients propagate through them, but the generator objective only updates its trainable dedicated parameters.

Figure \ref{fig:overall_architecture} illustrates the comprehensive pipeline of the proposed StaX-Net framework, which distinguishes itself by explicitly interleaving task-specific modules with multi-scale shared computational blocks across both the generator and discriminator. To ensure rich feature propagation, the downsampled features are further processed and merged with the preceding spatial transformations before advancing to the next shared stage. Crucially, multi-scale target MRI slices guide FAST layers exclusively during training, acting as structural priors that are entirely omitted at inference. The generative pathway then reconstructs the MRI representation by progressively upsampling the bottleneck features. 

At the same time, the discriminator evaluates the images through a sequential multi-scale pathway. This pathway alternates between its own dedicated layers and the exact same shared blocks that the generator uses. Finally, a fully connected network performs the final classification. Functionally, this structural parameter sharing enforces semantic feature alignment across two domains. It provides the necessary architectural foundation for the subsequent Stackelberg-inspired asymmetric optimization.

The coupled system is trained through separate leader ($D$) and follower ($G$) objectives. The follower composite loss is:
\begin{equation}
\begin{split}
\mathcal{L}_G(\theta_{G_{\text{ded}}}, \theta_{\text{shared}}) &= \lambda_{\text{adv}} \mathcal{L}_{\text{adv}}(G, D) \\
&\quad + \lambda_{L_{1}} \mathcal{L}_{L_{1}}(y^*, \hat{y}) \\
&\quad + \lambda_{\text{SSIM}} \mathcal{L}_{\text{SSIM}}(y^*, \hat{y}) \\
&\quad + \sum_{l=1}^{L} \mathcal{L}_{\text{FAST}}^{(l)}
\end{split}
\label{eq:generator_loss}
\end{equation}
where $\mathcal{L}_{\text{adv}}$ is the adversarial loss, $\mathcal{L}_{L_{1}}$ provides pixel-level geometric anchoring, $\mathcal{L}_{\text{SSIM}}$ preserves structural similarity, and $\lambda_{\text{adv}}, \lambda_{L_{1}}, \lambda_{\text{SSIM}} > 0$ are the balancing hyperparameters. Furthermore, $L$ denotes the total number of FAST layers integrated across different spatial scales within the generator, and $\mathcal{L}_{FAST}^{(l)}$ represents the target-supervised prior distillation loss computed at the $l$-th FAST layer. The leader loss $\mathcal{L}_D(\theta_{\text{shared}}, \theta_D)$ is formulated using a mean squared error (MSE) objective and optimized independently during the leader phase.
\subsection{Stackelberg-Inspired Asymmetric Optimization}
\subsubsection{Parameter Space Decomposition}
In standard GANs, the parameter spaces are completely disjoint ($\theta_G \cap \theta_D = \emptyset$). In StaX-Net, the $K$ blocks participate in both forward pathways but require a single, well-defined update rule to prevent conflicting gradients. If $\theta_{\text{shared}}$ were updated simultaneously by both $\mathcal{L}_G$ and $\mathcal{L}_D$, the opposing objectives could drive the shared representation in competing directions, causing instability. To resolve this, we assign explicit asymmetric parameter ownership. We define the leader's strategy space as $\Omega_D = \{\theta_{\text{shared}}, \theta_{D_{\text{ded}}}\}$ and the follower's strategy space as $\Omega_G = \{\theta_{G_{\text{ded}}}, \theta_{\text{FAST}}\}$.

\subsubsection{Sequential Leader-Follower Dynamics}
A rigorous Stackelberg formulation requires analytical hypergradients for the exact best-response, which are computationally prohibitive for deep networks \cite{fiez2020implicit}. Instead, we approximate leader-follower dynamics through a sequential optimization scheme stabilized by asymmetric parameter freezing and asymmetric learning rates ($\eta_G > \eta_D$). 

At iteration $t$, the discriminator acts as the leader, updating both $\theta_{\text{shared}}$ and $\theta_{D_{\text{ded}}}$ via $\mathcal{L}_D$ to establish the optimal feature representation. Subsequently, the generator acts as the follower. Notably, since $\theta_{\text{shared}} \in \Omega_D$, the shared parameters are explicitly frozen. The generator updates only its dedicated parameters $\Omega_G$ via $\mathcal{L}_G$. It adapts its dedicated synthesis pathway to the representation established by the preceding discriminator-controlled update. This removes a primary source of gradient conflict and provides a tractable Stackelberg-inspired optimization (detailed in Algorithm \ref{alg:stax_training}).

\subsection{Feature Aggregation and Spatial Transformation Layer}

\begin{figure}[htbp]
    \centering
    \includegraphics[trim={3.5cm 2cm 4.5cm 2cm}, clip, width=0.7\columnwidth]{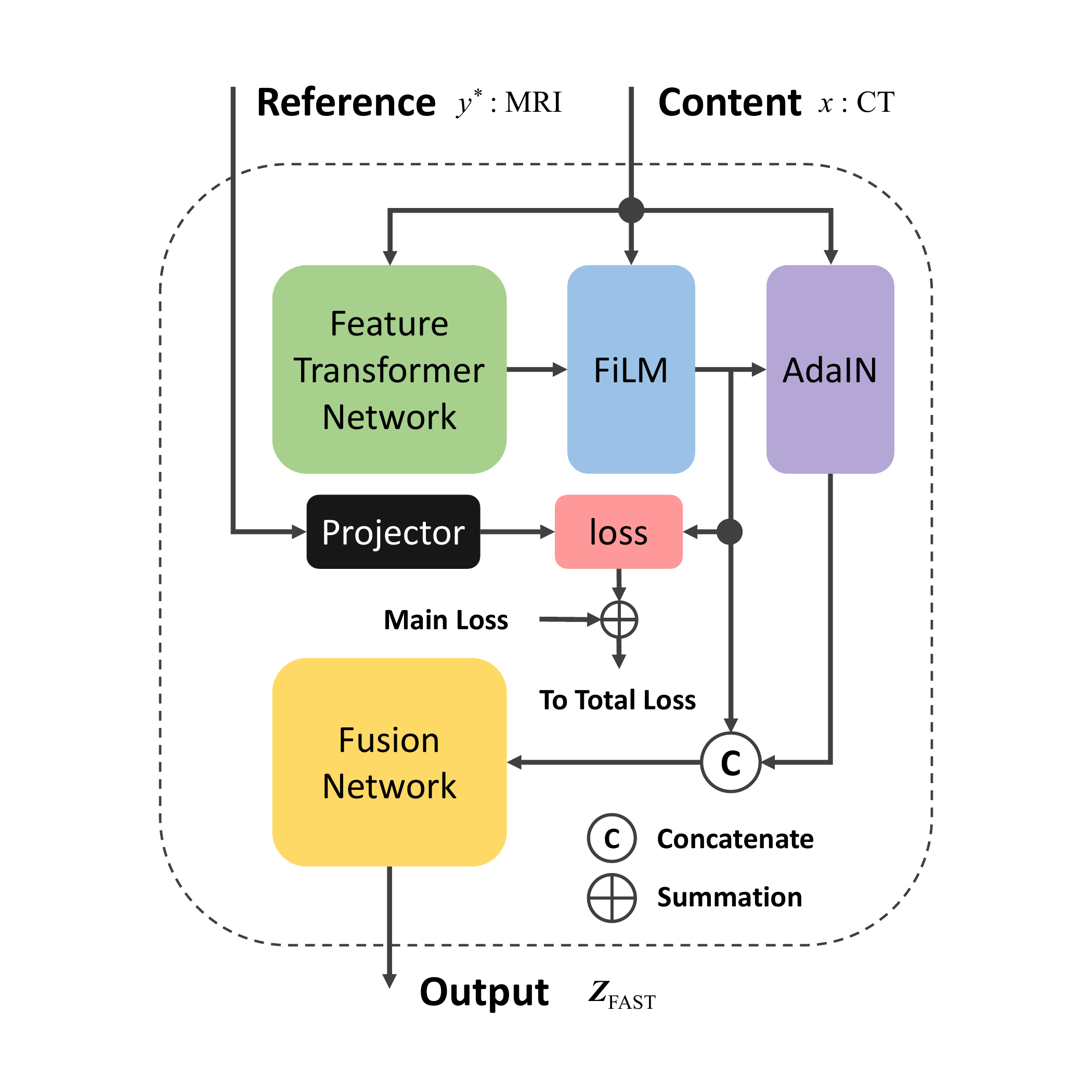}
    \caption{Detailed architecture of the FAST layer. The internal loss ($\mathcal{L}_{\mathrm{FAST}}$ in \ref{eq:fast_loss}) computed by this module during the forward pass is injected into the total optimization objective.}
    \label{fig:fast_layer}
\end{figure}

\subsubsection{Formal Notation and LUPI Paradigm}

Within this asymmetric framework, we introduce the FAST layer. We denote by $x \in \mathcal{X}$ a source-domain 2D CT slice and by $y^* \in \mathcal{Y}$ the co-registered target MRI slice. Parameterized by $\theta_{\text{FAST}}$, this layer acts as a feature transformer network that implements a source-only forward mapping (detailed in Fig. \ref{fig:fast_layer}):
\begin{equation}
F_{\theta_{\text{FAST}}}: \mathcal{X} \rightarrow \mathcal{F}_{\text{target}},
\label{eq:fast_mapping}
\end{equation}
where $\mathcal{F}_{\text{target}}$ represents a target-informed surrogate feature space. Under the LUPI paradigm~\cite{chen2022beyond, ahmed2021fame}, the privileged target MRI (reference) is never injected into the forward pass. Instead, it is utilized exclusively as a supervisory signal to compute the FAST loss, thereby distilling its structural representations into the transformed content features. During inference, the target is completely absent, and the FAST layer operates strictly as a deterministic source-only transformation.

The implementation constructs the dynamic prior bank as an intermediate feature map rather than as an external memory repository. Let $\mathbf{C}_{e}=\phi_{\mathrm{exp}}(x)$ denote the expanded content representation produced by the FAST layer. A spatial prior is then extracted by a 2D convolutional subnetwork containing standard and dilated convolutional branches, followed by a squeeze-and-excitation (SE) module:
\begin{equation}
\mathbf{P}=\operatorname{SE}\!\left(\psi_{\mathrm{sp}}^{\mathrm{dil}}(\mathbf{C}_{e})\right),
\label{eq:spatial_prior}
\end{equation}
where $\psi_{\mathrm{sp}}^{\mathrm{dil}}$ denotes the spatial convolution operator with a dilated branch. Channel-wise content statistics are computed from the content stream $\mathbf{C}_{e}$,
\begin{equation}
[\boldsymbol{\mu}_{C_{e}},\boldsymbol{\sigma}_{C_{e}}]=\operatorname{Stats}(\mathbf{C}_{e}),
\end{equation}
where $\operatorname{Stats}(\cdot)$ returns the channel-wise mean and standard deviation across spatial locations. These content-derived statistics are mapped to feature-wise linear modulation (FiLM) parameters by pointwise convolutions:
\begin{equation}
\boldsymbol{\gamma}=\kappa_{\gamma}\!\left(
[\boldsymbol{\mu}_{C_{e}};\boldsymbol{\sigma}_{C_{e}}]\right),\qquad
\boldsymbol{\beta}=\kappa_{\beta}\!\left(
[\boldsymbol{\mu}_{C_{e}};\boldsymbol{\sigma}_{C_{e}}]\right),
\label{eq:film}
\end{equation}
where $\kappa_{\gamma}$ and $\kappa_{\beta}$ are point-wise convolutional mappings. The dynamic prior bank is then formed by FiLM modulation of the spatial prior:
\begin{equation}
\mathbf{B}=\boldsymbol{\gamma}\odot\mathbf{P}+\boldsymbol{\beta},
\label{eq:dynamic_prior_bank}
\end{equation}
where $\mathbf{B}$ corresponds to the prior bank. Importantly, the modulation statistics are computed from the CT content stream, not from the privileged MRI.

The dynamic prior bank subsequently provides the affine statistics used to style the original content representation via adaptive instance normalization (AdaIN):
\begin{equation}
\widehat{\mathbf{C}}_{e}
=
\boldsymbol{\sigma}_{B}\odot
\frac{\mathbf{C}_{e}-\boldsymbol{\mu}_{C_{e}}}
{\boldsymbol{\sigma}_{C_{e}}}
+\boldsymbol{\mu}_{B},
\label{eq:adain}
\end{equation}
where $[\boldsymbol{\mu}_{B},\boldsymbol{\sigma}_{B}]=\operatorname{Stats}(\mathbf{B})$. Finally, the styled content $\widehat{\mathbf{C}}_{e}$ and the dynamic prior bank $\mathbf{B}$ are concatenated along the channel dimension (denoted by $\parallel$) and fed into the fusion subnetwork $\psi_{\mathrm{fuse}}(\cdot)$ to yield the final output feature map $\mathbf{Z}_{\mathrm{FAST}}$ of the FAST layer:
\begin{equation}
\mathbf{Z}_{\mathrm{FAST}}
=\psi_{\mathrm{fuse}}\!\left([\widehat{\mathbf{C}}_{e}\parallel\mathbf{B}]\right).
\label{eq:fast_output}
\end{equation}
Therefore, the implemented FAST layer processes the input through a sequential pipeline starting with content expansion. This is followed by spatial convolution and SE prior extraction. The extracted features then undergo content-driven FiLM modulation and prior-driven AdaIN styling, ultimately culminating in a final fusion step.

\subsubsection{Phase 1 - Target-Supervised Prior Distillation}
During training, the paired target MRI is available. However, the privileged target is never concatenated to the forward input. Instead, the FAST layer receives only the content input $x$ (representing the source CT slice, as illustrated in Fig. \ref{fig:fast_layer}) and executes the source-only forward mapping $F_{\theta_{\mathrm{FAST}}}(x)$. The privileged target $y^*$ serves as the reference input, which is processed by a pathway-specific projection block $\Phi$ (the \textit{Projector} in Fig. \ref{fig:fast_layer}) to obtain the projected target feature $\mathbf{Y} = \Phi(y^*)$. 

This projection aligns the raw reference MRI to the spatial resolution and channel dimensionality of the dynamic prior bank $\mathbf{B}$. Finally, both $\mathbf{B}$ and $\mathbf{Y}$ are utilized exclusively within the auxiliary loss $\mathcal{L}_{\mathrm{FAST}}$ to guide the spatial priors in a shared latent space.

By comparing representations in the feature space rather than the raw pixel space, the FAST loss effectively distills target-domain structural characteristics into the source-derived prior. This loss couples a feature-level intensity anchor with a structural-alignment branch:
\begin{equation}
\mathcal{L}_{\mathrm{L_{1}}} = \left\|\mathbf{B}-\mathbf{Y}\right\|_{1},
\end{equation}
\begin{equation}
\begin{split}
\mathcal{L}_{\mathrm{grad}} &= \left\|\nabla_{u}\mathbf{B}-\nabla_{u}\mathbf{Y}\right\|_{1} + \left\|\nabla_{v}\mathbf{B}-\nabla_{v}\mathbf{Y}\right\|_{1}.
\end{split}
\end{equation}
where $u$ and $v$ are the coordinates of the image and $\left\| \cdot \right\|_{1}$ referes to $L_{1}$-norm. The implemented auxiliary FAST loss is therefore,
\begin{equation}
\begin{split}
\mathcal{L}_{\mathrm{FAST}} &= \lambda_{\mathrm{L_{1}}}\mathcal{L}_{\mathrm{L_{1}}} + \lambda_{\mathrm{grad}}\mathcal{L}_{\mathrm{grad}}.
\end{split}
\label{eq:fast_loss}
\end{equation}
where $\lambda_{\mathrm{L_{1}}}, \lambda_{\mathrm{grad}} > 0$. During the generator-controlled update phase, $\mathcal{L}_{\mathrm{FAST}}$ acts as a structural regularizer and is directly added to the generator's total objective $\mathcal{L}_G$ (Eq. \ref{eq:generator_loss}). Consequently, the parameters $\theta_{\mathrm{FAST}}$ are updated jointly with the dedicated generator parameters $\theta_{G_{\text{ded}}}$ via standard backpropagation, avoiding the need for an independent optimization stage.

\subsubsection{Phase 2 - Inference without Target Supervision}

At inference, the model receives the source input alone, while the target modality is completely omitted. The forward pass relies exclusively on the deterministic source-only transformation $F_{\theta_{\mathrm{FAST}}^{*}}(x)$, where the FAST layer receives only the content input (the source CT slice $x$) and utilizes the optimized parameters $\theta_{\mathrm{FAST}}^{*}$ learned during the target-supervised training phase. 

Because the dynamic prior bank $\mathbf{B}$ and content representations are generated on the fly directly from $x$ using the frozen internal mappings, the FAST layer applies the structural priors acquired during training without requiring any target-dependent adaptation.

\begin{algorithm}[htbp]
\caption{Stackelberg-Inspired Asymmetric Training for StaX-Net}
\label{alg:stax_training}
\begin{algorithmic}[1]
\REQUIRE Co-registered Target MRI $y^*$, Source CT $x$, Loss weights $\lambda_{\text{adv}}, \lambda_{L_{1}}, \lambda_{\text{SSIM}}$
\REQUIRE Leader parameters: $\Omega_D = \{\theta_{\text{shared}}, \theta_{D_{\text{ded}}}\}$
\REQUIRE Follower parameters: $\Omega_G = \{\theta_{G_{\text{ded}}}, \theta_{\text{FAST}}\}$
\REQUIRE Asymmetric learning rates $\eta_D, \eta_G$ (where $\eta_G > \eta_D$)
\WHILE{not converged}
    \STATE \textbf{// Phase 1: Leader (Discriminator) Update}
    \STATE Enable updates for $\Omega_D$; hold $\Omega_G$ fixed
    \STATE Sample mini-batch of source $x$ and target $y^*$
    \STATE Evaluate generator forward pathway to synthesize fake images $\hat{y}$
    \STATE Compute $\mathcal{L}_D$ on real $y^*$ and fake $\hat{y}$
    \STATE Update $\Omega_D$ using $\nabla_{\Omega_D} \mathcal{L}_D$ with rate $\eta_D$
    
    \STATE \textbf{// Phase 2: Follower (Generator) Update}
    \STATE Hold $\Omega_D$ fixed; enable updates for $\Omega_G$
    \STATE Evaluate generator forward pathway using the fixed updated $\theta_{\text{shared}}$
    \STATE Compute $\mathcal{L}_G$ via Eq. \eqref{eq:generator_loss} (including FAST loss)
    \STATE Update $\Omega_G$ using $\nabla_{\Omega_G} \mathcal{L}_G$ with rate $\eta_G$
\ENDWHILE
\end{algorithmic}
\end{algorithm}

\subsection{Implicit Feature Matching}
By forcing $G$ to utilize the $D$'s shared feature extractor, implicit feature matching becomes an architectural constraint rather than a separate statistical penalty \cite{liu2016coupled}. While standard feature matching explicitly minimizes the statistical distance between the expected features of real and generated data within the discriminator's hidden layers \cite{salimans2016improved}, our proposed architecture enforces this alignment structurally: the $G$ decoder blocks receive features from the same shared computational graph, convolution weights, and nonlinear activations used by $D$ \cite{sendik2019crossnet}.

\subsection{Cooperative-Competitive Hybrid Loss}
Clinical synthesis should transfer source anatomy rather than invent structures to satisfy adversarial feedback. The follower loss therefore combines an adversarial term with deterministic constraints: $L_{1}$ anchors global anatomy, SSIM preserves local structure, and the adversarial loss encourages target-like high-frequency texture. This combination of complementary loss functions ensures balanced optimization and precise boundary delineation \cite{golnari2026adaptive}, constraining the synthesis toward the paired target.

\section{Theoretical Analysis}

\subsection{Target-Informed Feature Distillation}
In a standard conditional GAN, the optimal generated output set is determined primarily by the conventional adversarial minimax objective. However, this optimization is not explicitly constrained by a target-supervised prior in the latent space. In the proposed framework, the FAST transform incorporates privileged MRI information during training. Furthermore, because the shared computational blocks ($\theta_{\text{shared}}^*$), with the superscript $^*$ denoting optimal parameters obtained upon training completion, are explicitly frozen during the generator's update phase, we can interpret the generated set as being conditioned on both a fixed leader strategy and a structurally regularized feature mapping:
\begin{equation}
\begin{split}
\mathcal{S}_{\text{FAST}} &= \Big\{ G_{\theta_{G_{\text{ded}}}^*}\!\left(F_{\theta_{\text{FAST}}^{*}}(x)\right) \;\Big\vert\; \\
&\quad \theta_{G_{\text{ded}}}^*, \theta_{\text{FAST}}^{*} \text{ obtained jointly given fixed } \theta_{\text{shared}}^* \\
&\quad \text{under } \mathcal{L}_G + \mathcal{L}_{\mathrm{FAST}}(x,y^*) \Big\}.
\end{split}
\end{equation}

Since $\theta_{\text{FAST}}^{*}$ is optimized jointly with the generator's dedicated parameters using feature-level target-domain supervision, while adapting to the fixed shared representation, $\mathcal{S}_{\text{FAST}}$ is heuristically guided by a target-informed feature transformation.While this does not strictly constrain the output to a mathematical subspace, it introduces a strong inductive bias. By acting as a structural regularizer, the LUPI-driven FAST layer encourages the generator to prioritize representations that are anatomically consistent with the target modality, thereby mitigating the risk of unconstrained adversarial hallucinations as supported by our empirical evaluations.

\subsection{Optimization Stability via Asymmetric Updates}
In conventional coupled architectures, simultaneous updating of the shared parameters by opposing generator and discriminator objectives often induce severe gradient conflicts. The proposed asymmetric parameter ownership explicitly mitigates this issue. By enforcing a strict leader-follower update schedule, the shared manifold acts as a cooperative-competitive constraint \cite{liu2016coupled, schaffer2019implicit} without subjecting the shared weights to adversarial gradient collisions. 

Selective freezing of the shared blocks during the generator's phase ensures that the follower must adapt to the representation established by the leader, rather than disrupting it. While we do not claim formal analytical convergence to a strict Stackelberg equilibrium, which would require computationally prohibitive hypergradient evaluations \cite{fiez2020implicit}. Combined with deterministic $L_{1}$ and SSIM anchors, this asymmetric optimization heuristic effectively reduces training volatility and limits feature-space divergence.

\subsection{Design Justification}
The decision to freeze only $\theta_{\text{shared}}$ during the generator update strikes a critical balance between the representational stability and the generative capacity. If the entire discriminator were frozen, the adversarial signal would stagnate; if the shared blocks were trainable by the generator, the target-domain feature alignment would be compromised. Consequently, updating only the trainable dedicated blocks $\theta_{G_{\text{ded}}}$ allows the generator to effectively map the fixed shared features to the target domain. The use of asymmetric learning rates ($\eta_G > \eta_D$) further compensates for the restricted update space of the generator, allowing the follower to adapt faster to the leader's strategy without requiring exact multi-step unrolling or formal best-response calculations \cite{fiez2020implicit}.

\section{Experiments}

\subsection{Dataset and Preprocessing}
In all the experiments here we use the SynthRAD 2025 Grand Challenge dataset \cite{thummerer2025synthrad2025}, including the Abdomen/Pelvis, Head and Neck, and Thorax subsets. Each subset provides paired CT and MRI volumes. The data are partitioned at the subject level into 80\% training, 10\% validation, and 10\% testing to prevent slice-level leakage. To prevent uninformative data from corrupting, both CT and MRI volumes are truncated along the z-axis to remove empty or low-tissue slices at the volume boundaries, while fully preserving the intermediate slice sequence. The in-plane dimensions are then resampled to a $256 \times 256$ lattice (using linear interpolation for CT and B-spline for MRI) while retaining the original z-axis depth. Finally, a 3D volumetric global normalization is applied to map intensities to the $[-1, 1]$ range using modality-specific clipping thresholds ($[-3000, 3000]$ Hounsfield units (HU) for CT, and the $0.5^{\text{th}}$ to $99.5^{\text{th}}$ percentile for MRI). During training, geometric augmentations applied include random $90^\circ$ rotations, spatial shifting (translations), and horizontal/vertical flips.

\subsection{Implementation Details and Baselines}
We compare StaX-Net against five baselines, including Pix2Pix \cite{isola2017image}, Vanilla cGAN \cite{mirza2014conditional}, UNet-AE \cite{ronneberger2015u}, CycleGAN \cite{zhu2017unpaired}, and an adapted MedViT-cGAN \cite{manzari2023medvit}. Because the original MedViT architecture was formulated for generalized medical image classification, we adapted its robust representation learning capabilities to the image translation task. Specifically, we utilized the MedViT efficient self-attention blocks, alongside patch merging and expanding modules, to construct a hierarchical U-Net style encoder-decoder generator. This transformer-based generator was then paired with a convolutional PatchGAN \cite{isola2017image} discriminator.

All models were trained over 150 epochs with a batch size of 32. The Adam optimizer \cite{kingma2014adam} was employed for trainable parameters with a first-order momentum coefficient of $\beta_1 = 0.5$ and the default second-order momentum of $\beta_2 = 0.999$. Asymmetric learning rates were employed as an empirical stabilization heuristic: the $G$ learning rate was set to $1.25 \times 10^{-4}$, and the $D$ learning rate was set to $1 \times 10^{-4}$.

To stabilize the framework optimization, label smoothing was applied to the $D$'s targets: real labels were smoothed using $0.9$ (range $[0.9,1]$), and fake labels using $0.1$ (range $[0,0.1]$). The adversarial loss was optimized using a mean squared error (MSE) formulation rather than the conventional binary cross-entropy. The follower loss in Eq.~\eqref{eq:generator_loss} used $\lambda_{\text{adv}} = 1.0$.

The FAST layer's dual-objective loss (Eq.~\eqref{eq:fast_loss}) used balanced weights $\lambda_{\mathrm{L_{1}}} = 0.5$ and $\lambda_{\mathrm{grad}} = 0.5$ for target-supervised prior distillation. The reconstruction loss was formulated as:
\begin{equation}
\mathcal{L}_{recon}
=
\lambda_{L_{1}}\,\mathcal{L}_{L_{1}}
+
\lambda_{\text{SSIM}}\,(1-\mathrm{SSIM}),
\label{eq:recon_loss}
\end{equation}
where the balancing weights were set to $\lambda_{L_{1}}, \lambda_{\text{SSIM}} > 0$, and SSIM was computed assuming a maximum signal dynamic range of $2.0$ to match the $[-1,1]$ intensity range.

\subsection{Evaluation Metrics}
Quantitative evaluation was performed using a five-metric protocol spanning classical fidelity, deep perceptual distance, distribution alignment, and frequency-domain realism:

\begin{itemize}
\item \textbf{Classical Fidelity (PSNR and SSIM):} Peak signal-to-noise ratio (PSNR) evaluates global pixel-level error, while SSIM \cite{wang2004image} measures local luminance, contrast, and structural preservation.

\item \textbf{Deep Perceptual Feature Distance:} To evaluate structural and textural consistency beyond simple pixel-level differences, we extract deep features using a VGG16 network \cite{simonyan2014very} pre-trained on ImageNet. While ImageNet semantic classes are distinct from medical anatomies, early and intermediate CNN layers (truncated at the third convolutional layer of the third block) act as universal, domain-agnostic edge and texture extractors that strongly correlate with perceptual structural fidelity \cite{zhang2018unreasonable}. Operating slice-wise, the 2D grayscale images are triplicated across the channel dimension, scaled, and normalized to the VGG expected input space. Following feature extraction, global average pooling is applied to yield a compact, spatially-invariant semantic descriptor $\phi_{\text{vgg}}(\cdot) \in \mathbb{R}^C$ (where $C=256$). For a synthesized 2D slice $\hat{y}$ and its paired ground truth $y^*$, the pairwise perceptual distance $\mathcal{L}_{\text{VGG-MSE}}$ is defined as the mean squared error between these pooled descriptors:
\begin{equation}
\mathcal{L}_{\text{VGG-MSE}} = \frac{1}{C} \sum_{c=1}^{C} \left( \phi_{\text{vgg}}(\hat{y})_c - \phi_{\text{vgg}}(y^*)_c \right)^2
\end{equation}
where $c$ indexes the feature channels. Lower values indicate closer perceptual and textural correspondence.

\item \textbf{Distribution Alignment via Fréchet Feature Distance:} To evaluate whether the generator models the target feature distribution, we measure the Fréchet feature distance (FFD) \cite{heusel2017gans} between the multivariate Gaussian distributions fitted to the deep features of the real ($\mathcal{N}(\mu_r, \Sigma_r)$) and synthesized ($\mathcal{N}(\mu_g, \Sigma_g)$) test sets:
\begin{equation}
\begin{split}
\text{FFD} &= \left\| \mu_r - \mu_g \right\|_2^2 + \text{Tr}\left(\Sigma_r + \Sigma_g - 2(\Sigma_r \Sigma_g)^{1/2}\right)
\end{split}
\end{equation}
where $\mu_r$ and $\Sigma_r$ denote the mean vector and covariance matrix of the deep feature representations extracted from the real target images, while $\mu_g$ and $\Sigma_g$ represent the corresponding mean vector and covariance matrix for the synthesized (generated) images. Additionally, $\left\| \cdot \right\|_2^2$ represents the squared Euclidean distance between the mean vectors, and $\text{Tr}(\cdot)$ denotes the matrix trace operator. A lower FFD indicates closer feature-distribution alignment.

\item \textbf{Frequency Domain Analysis:} To quantify the preservation of global frequency content and fine structural textures, the radially averaged power spectrum is computed for real and synthesized images using 2D fast Fourier transform (FFT).

\item \textbf{Volumetric Consistency and 3D Metrics:} To assess cross-slice continuity and longitudinal smoothness, we compute the z-axis gradient mean absolute error (Z-MAE). Z-MAE is defined as the mean absolute difference between the longitudinal gradients (differences between adjacent slices along the z-axis) of the synthesized volume and the ground truth. Furthermore, all reported 3D metrics (including 3D PSNR, SSIM, MAE, and RMSE) are calculated directly from the fully reconstructed 3D volumes rather than being averaged slice-by-slice.
\end{itemize}

\subsection{Results and Evaluation}
We evaluated the proposed StaX-Net against representative baselines (Pix2Pix, Vanilla cGAN, UNet-AE, CycleGAN, MedViT) across three anatomical subsets (Abdomen/Pelvis, Thorax, and Head \& Neck). Table~\ref{tab:quantitative_results_combined} summarizes the quantitative metrics for each modality, including mean absolute error (MAE) in 2D and 3D, and root mean square error (RMSE) in 3D. Furthermore, we contextualize our volumetric performance against MD-DGA \cite{ni2026medical}, a recently proposed 3D diffusion framework evaluated on the earlier SynthRAD 2023 dataset. Despite relying entirely on an efficient 2D architecture, StaX-Net achieves superior volumetric fidelity compared to this computationally intensive 3D baseline.

\begin{table*}[htbp]
\caption{Comprehensive quantitative comparison across the SynthRAD 2025 Abdomen/Pelvis (AB), Head \& Neck (HN), and Thorax (TH) datasets. Values are presented as mean (std).}
\begin{center}
\resizebox{\textwidth}{!}{
\begin{tabular}{@{}llcccccccccc@{}}
\toprule
\textbf{Data} & \textbf{Method}
& \textbf{PSNR (2D)} $\uparrow$
& \textbf{SSIM (2D)} $\uparrow$
& \textbf{MAE (2D)} $\downarrow$
& \textbf{VGG-MSE} $\downarrow$
& \textbf{FFD} $\downarrow$
& \textbf{PSNR (3D)} $\uparrow$
& \textbf{SSIM (3D)} $\uparrow$
& \textbf{MAE (3D)} $\downarrow$
& \textbf{RMSE (3D)} $\downarrow$
& \textbf{Z-MAE} $\downarrow$
\\
\midrule
\multirow{6}{*}{\textbf{AB}} 
& UNet-AE & 17.71 (2.29) & 0.674 (0.102) & 0.1281 (0.0502) & 1687.9 (673.6) & 369,420 & 17.48 (1.87) & 0.653 (0.106) & 0.1299 (0.0462) & 0.2737 (0.0597) & 0.0460 (0.0133) \\
& Vanilla cGAN & 15.87 (2.73) & 0.635 (0.102) & 0.1695 (0.0851) & 4626.8 (5702.4) & 1,033,639 & 15.41 (2.22) & 0.613 (0.096) & 0.1742 (0.0773) & 0.3520 (0.1057) & 0.0435 (0.0149) \\
& Pix2Pix & 17.06 (2.17) & 0.652 (0.098) & 0.1371 (0.0499) & 991.2 (318.5) & 168,908 & 16.76 (1.65) & 0.632 (0.098) & 0.1394 (0.0435) & 0.2957 (0.0571) & 0.0522 (0.0139) \\
& CycleGAN & 14.28 (1.93) & 0.593 (0.107) & 0.2014 (0.0647) & 1074.5 (772.0) & 109,309 & 14.10 (1.37) & 0.575 (0.104) & 0.2021 (0.0564) & 0.3995 (0.0615) & 0.0832 (0.0175) \\
& MedViT & 17.14 (2.04) & 0.661 (0.100) & 0.1352 (0.0481) & 1164.8 (473.4) & 197,237 & 16.98 (1.78) & 0.635 (0.106) & 0.1366 (0.0441) & 0.2892 (0.0591) & 0.0643 (0.0201) \\
& \textbf{StaX-Net} & \textbf{17.37 (2.13)} & \textbf{0.669 (0.099)} & \textbf{0.1299 (0.0497)} & \textbf{642.1 (413.7)} & \textbf{86,534} & \textbf{17.14 (1.74)} & \textbf{0.651 (0.102)} & \textbf{0.1322 (0.0452)} & \textbf{0.2837 (0.0579)} & \textbf{0.0517 (0.0131)} \\
\midrule
\multirow{6}{*}{\textbf{HN}} 
& UNet-AE & 19.84 (1.95) & 0.674 (0.102) & 0.1731 (0.0504) & 1684.3 (955.8) & 339,202 & 19.19 (1.40) & 0.737 (0.026) & 0.0913 (0.0128) & 0.2224 (0.0350) & 0.0636 (0.0060) \\
& Vanilla cGAN & 15.52 (1.45) & 0.684 (0.067) & 0.2116 (0.0456) & 5438.7 (2495.5) & 1,224,454 & 15.28 (0.53) & 0.673 (0.026) & 0.1503 (0.0137) & 0.3452 (0.0212) & 0.0705 (0.0064) \\
& Pix2Pix & 18.47 (1.86) & 0.727 (0.058) & 0.1857 (0.0506) & 1247.1 (619.8) & 229,321 & 17.86 (1.41) & 0.720 (0.034) & 0.1065 (0.0181) & 0.2592 (0.0413) & 0.0713 (0.0074) \\
& CycleGAN & 15.36 (1.74) & 0.693 (0.059) & 0.2379 (0.0636) & 1500.2 (813.6) & 220,183 & 14.87 (0.85) & 0.687 (0.022) & 0.1571 (0.0203) & 0.3628 (0.0344) & 0.0877 (0.0096) \\
& MedViT & 18.89 (1.94) & 0.737 (0.060) & 0.2011 (0.0713) & 1188.5 (1049.1) & 190,903 & 18.18 (1.24) & 0.726 (0.029) & 0.1012 (0.0145) & 0.2491 (0.0343) & 0.0762 (0.0076) \\
& \textbf{StaX-Net} & \textbf{19.55 (2.02)} & \textbf{0.750 (0.058)} & \textbf{0.2008 (0.0632)} & \textbf{778.2 (553.6)} & \textbf{122,924} & \textbf{18.84 (1.49)} & \textbf{0.746 (0.032)} & \textbf{0.0915 (0.0153)} & \textbf{0.2318 (0.0384)} & \textbf{0.0695 (0.0087)} \\
\midrule
\multirow{6}{*}{\textbf{TH}} 
& UNet-AE & 19.26 (2.61) & 0.757 (0.072) & 0.1347 (0.0428) & 1462.5 (475.1) & 320,837 & 18.90 (2.09) & 0.746 (0.065) & 0.0926 (0.0276) & 0.2335 (0.0557) & 0.0365 (0.0092) \\
& Vanilla cGAN & 16.49 (2.13) & 0.713 (0.080) & 0.1664 (0.0438) & 2315.8 (826.6) & 488,840 & 16.17 (1.59) & 0.694 (0.073) & 0.1326 (0.0344) & 0.3158 (0.0556) & 0.0372 (0.0107) \\
& Pix2Pix & 18.44 (2.30) & 0.739 (0.070) & 0.1562 (0.0593) & 862.2 (407.2) & 142,601 & 18.14 (1.77) & 0.729 (0.063) & 0.1025 (0.0275) & 0.2530 (0.0520) & 0.0429 (0.0093) \\
& CycleGAN & 16.66 (1.89) & 0.707 (0.079) & 0.2045 (0.0657) & 537.3 (370.1) & 54,001 & 16.43 (1.44) & 0.690 (0.070) & 0.1318 (0.0339) & 0.3060 (0.0516) & 0.0615 (0.0117) \\
& MedViT & 18.93 (2.63) & 0.747 (0.076) & 0.1481 (0.0569) & 1061.4 (416.5) & 206,313 & 18.58 (2.29) & 0.732 (0.070) & 0.0982 (0.0297) & 0.2434 (0.0602) & 0.0453 (0.0116) \\
& \textbf{StaX-Net} & \textbf{18.67 (2.62)} & \textbf{0.748 (0.074)} & \textbf{0.1519 (0.0495)} & \textbf{612.5 (329.6)} & \textbf{88,961} & \textbf{18.37 (2.27)} & \textbf{0.738 (0.069)} & \textbf{0.1012 (0.0320)} & \textbf{0.2492 (0.0620)} & \textbf{0.0429 (0.0112)} \\
\bottomrule
\end{tabular}
}
\end{center}
\label{tab:quantitative_results_combined}
\end{table*}

\subsubsection{Classical Fidelity and Statistical Stability}

Across the evaluated anatomical subsets, StaX-Net provides competitive classical fidelity, with structural preservation reflected by SSIM and low absolute and squared errors. In the Thorax and Head \& Neck regions, StaX-Net achieves SSIM values of $0.748$ and $0.750$, respectively. StaX-Net also achieves MAE (3D) and RMSE (3D) values of $0.1012$ and $0.2492$ in Thorax and $0.0915$ and $0.2318$ in Head \& Neck, respectively, indicating close volumetric correspondence to the GT.

A critical observation emerges regarding the UNet-AE baseline. While UNet-AE records a higher PSNR value in the Head \& Neck dataset ($19.84$) and a competitive MAE score ($0.1281$ MAE (2D) in the AB dataset), deterministic architectures optimized primarily for reconstruction can favor smooth outputs. Such smoothing can reduce pixel-wise error while attenuating high-frequency micro-textures relevant to medical image interpretation.

\subsubsection{Qualitative Evaluation and Error Localization}

\begin{figure*}[htbp]
    \centering
    \resizebox{0.85\textwidth}{!}{
        \setlength{\tabcolsep}{3pt}
        \begin{tabular}{c c}
            
            $\vcenter{\hbox{\rotatebox{90}{Head \& Neck}}}$ & 
            $\vcenter{\hbox{\includegraphics[trim=0 0 0 0, clip, width=1.0\textwidth]{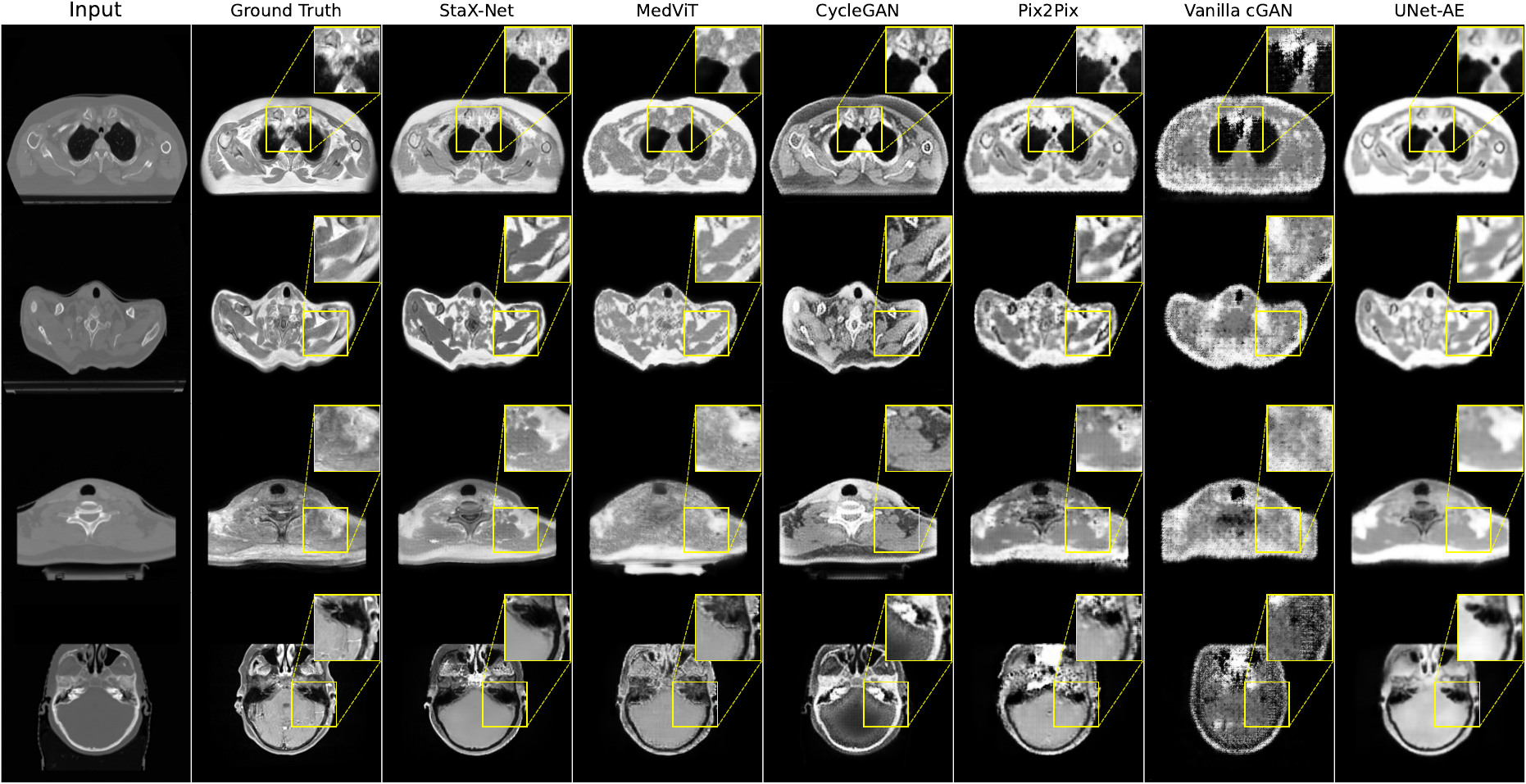}}}$ \\[0.4cm]
            
            $\vcenter{\hbox{\rotatebox{90}{Thorax}}}$ & 
            $\vcenter{\hbox{\includegraphics[trim=0 0 0 15, clip, width=1.0\textwidth]{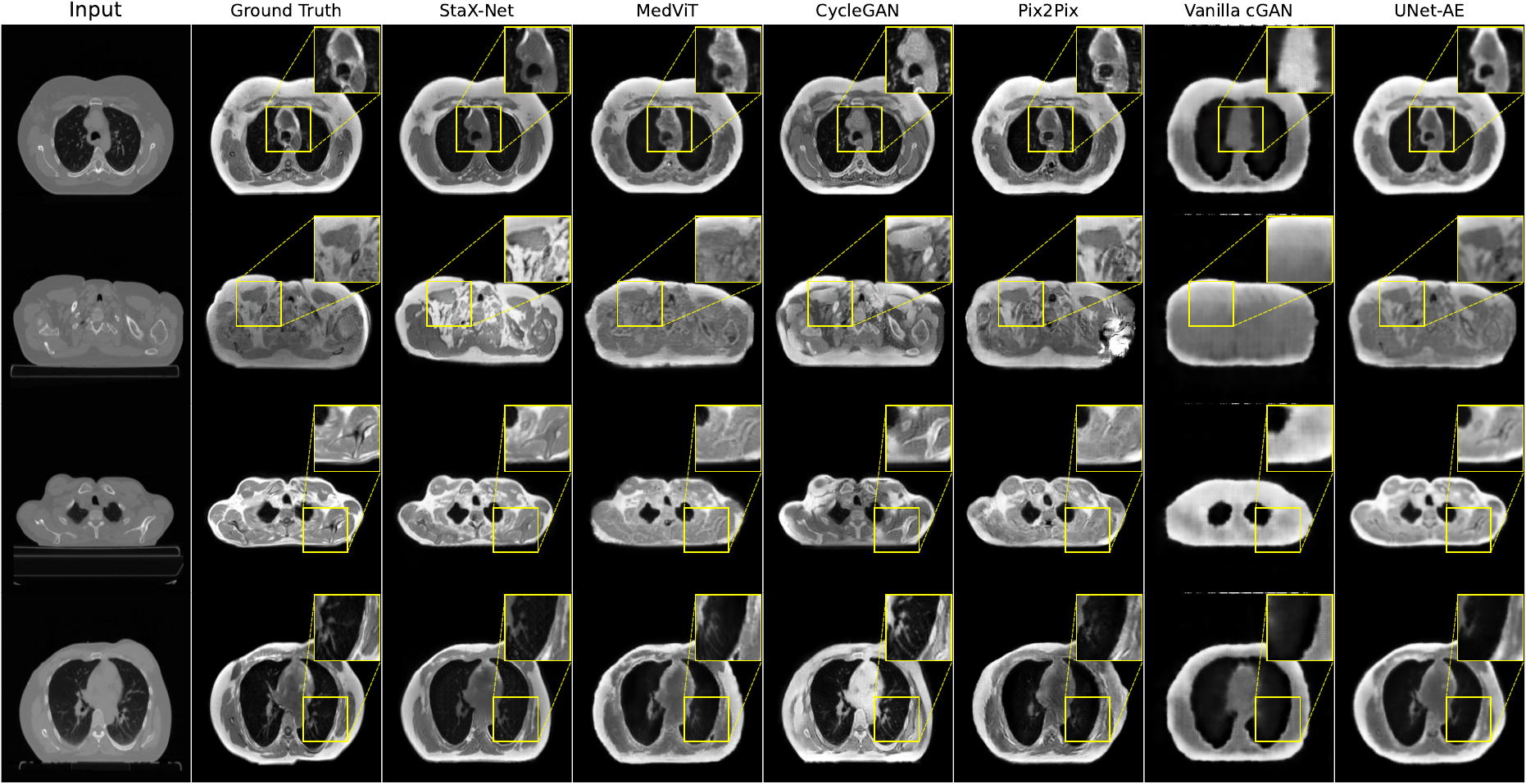}}}$ \\[0.4cm]
            
            $\vcenter{\hbox{\rotatebox{90}{Abdomen/Pelvis}}}$ & 
            $\vcenter{\hbox{\includegraphics[trim=0 0 0 15, clip, width=1.0\textwidth]{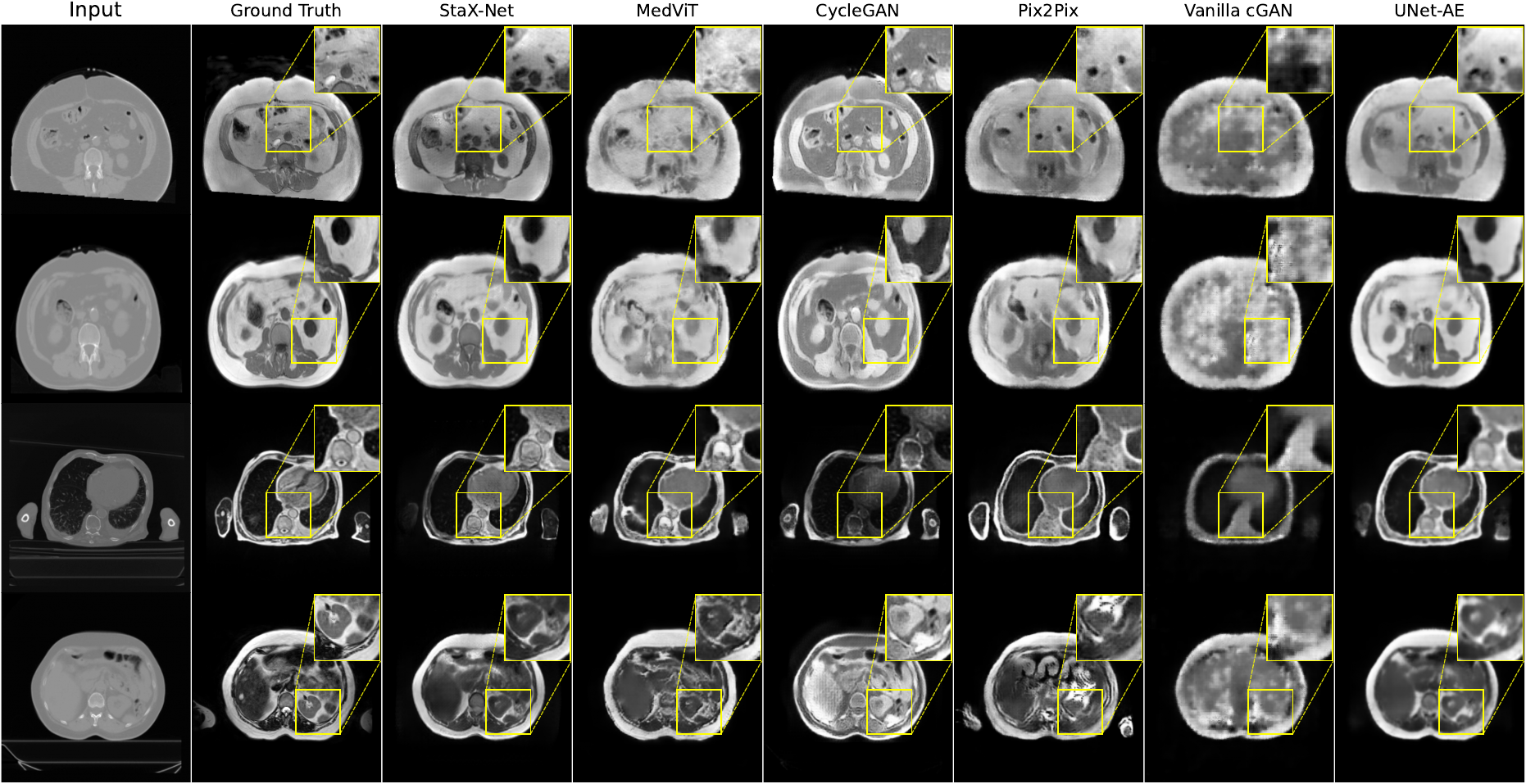}}}$ \\
            
        \end{tabular}
    }
    \caption{Qualitative comparison of visual fidelity and localized structural preservation. The rows display axial slices of different anatomical regions: Head \& Neck (top), Thorax (middle), and Abdomen/Pelvis (bottom). Columns from left to right show the input CT, the GT MRI, followed by the predicted MRI for StaX-Net, MedViT, CycleGAN, Pix2Pix, Vanilla cGAN, and UNet-AE. Zoomed-in insets (top-right) highlight local structures.}
    \label{fig:qualitative_predictions}
\end{figure*}

Visual fidelity was assessed via cross-modality synthesis grids (see Fig~\ref{fig:qualitative_predictions}). The zoomed-in regions of interest (ROIs) demonstrate the preservation of trabecular bone structures in the Head \& Neck region, soft-tissue boundaries in the Abdomen/Pelvis, and cardiac margins in the Thorax dataset.

Contrasting this with the baselines reveals different reconstruction characteristics. CycleGAN exhibits localized structural deviations associated with its unpaired mapping, while UNet-AE outputs smoother approximations.

\subsubsection{Frequency Domain Realism}

A signal-processing analysis was conducted using the radially averaged power spectrum via the 2D FFT. The GT spectrum provides a reference for the spatial-frequency characteristics of the target images.

\begin{figure*}[htbp]
    \centering
    \setlength{\tabcolsep}{2pt}
    \begin{tabular}{c c c}
        
        \includegraphics[trim=0 0 0 0, clip, width=0.32\textwidth]{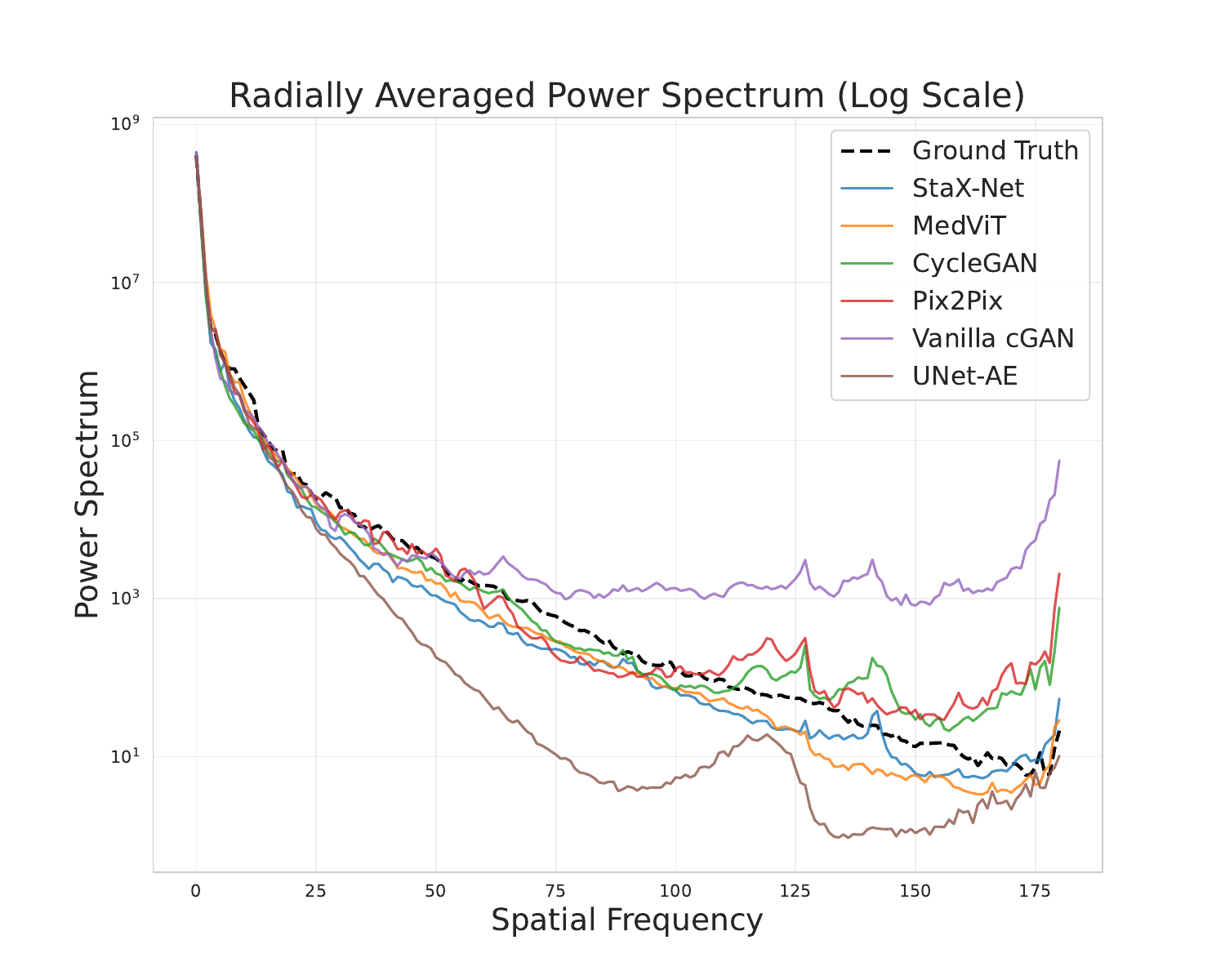} &
        \includegraphics[trim=0 0 0 0, clip, width=0.32\textwidth]{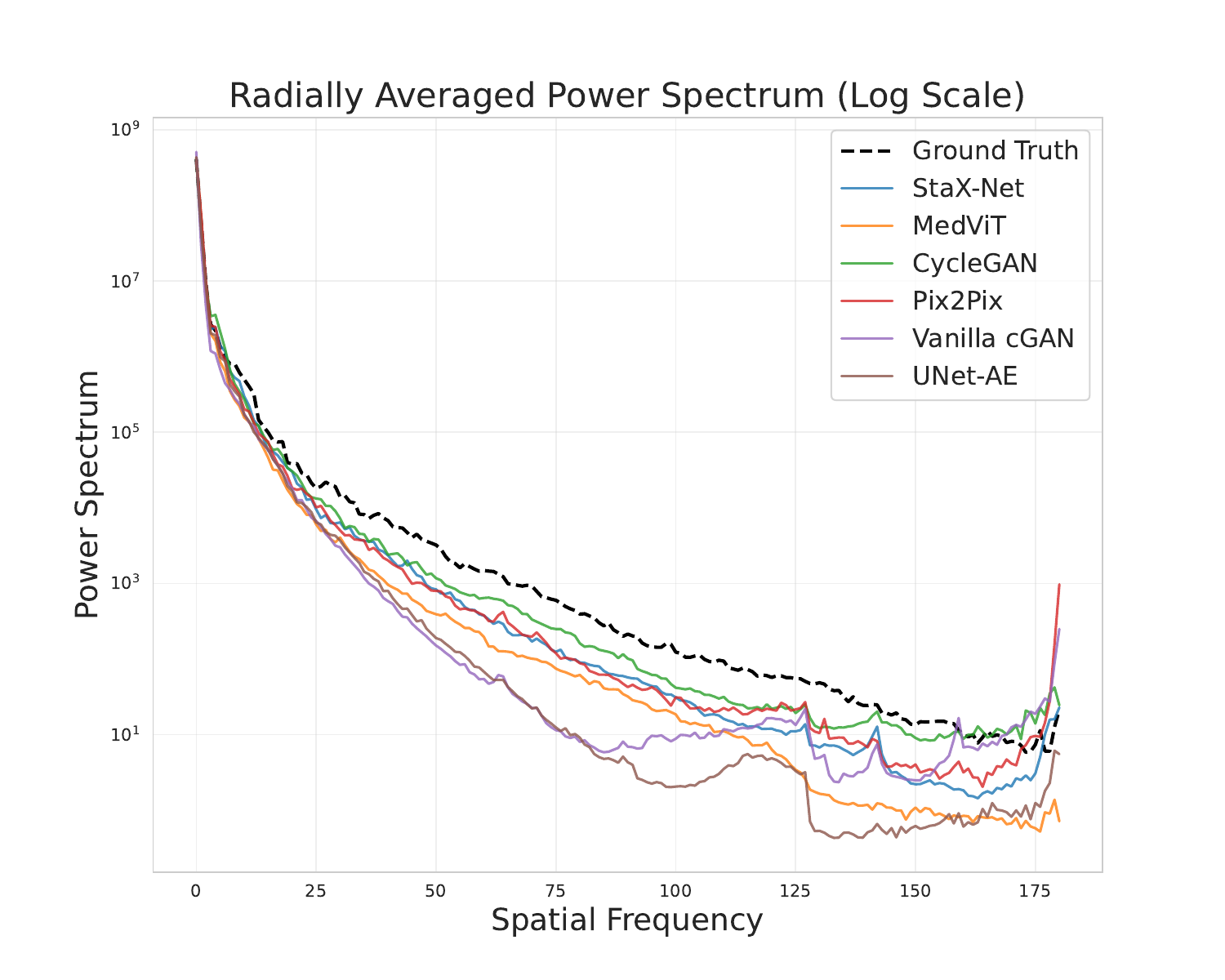} &
        \includegraphics[trim=0 0 0 0, clip, width=0.32\textwidth]{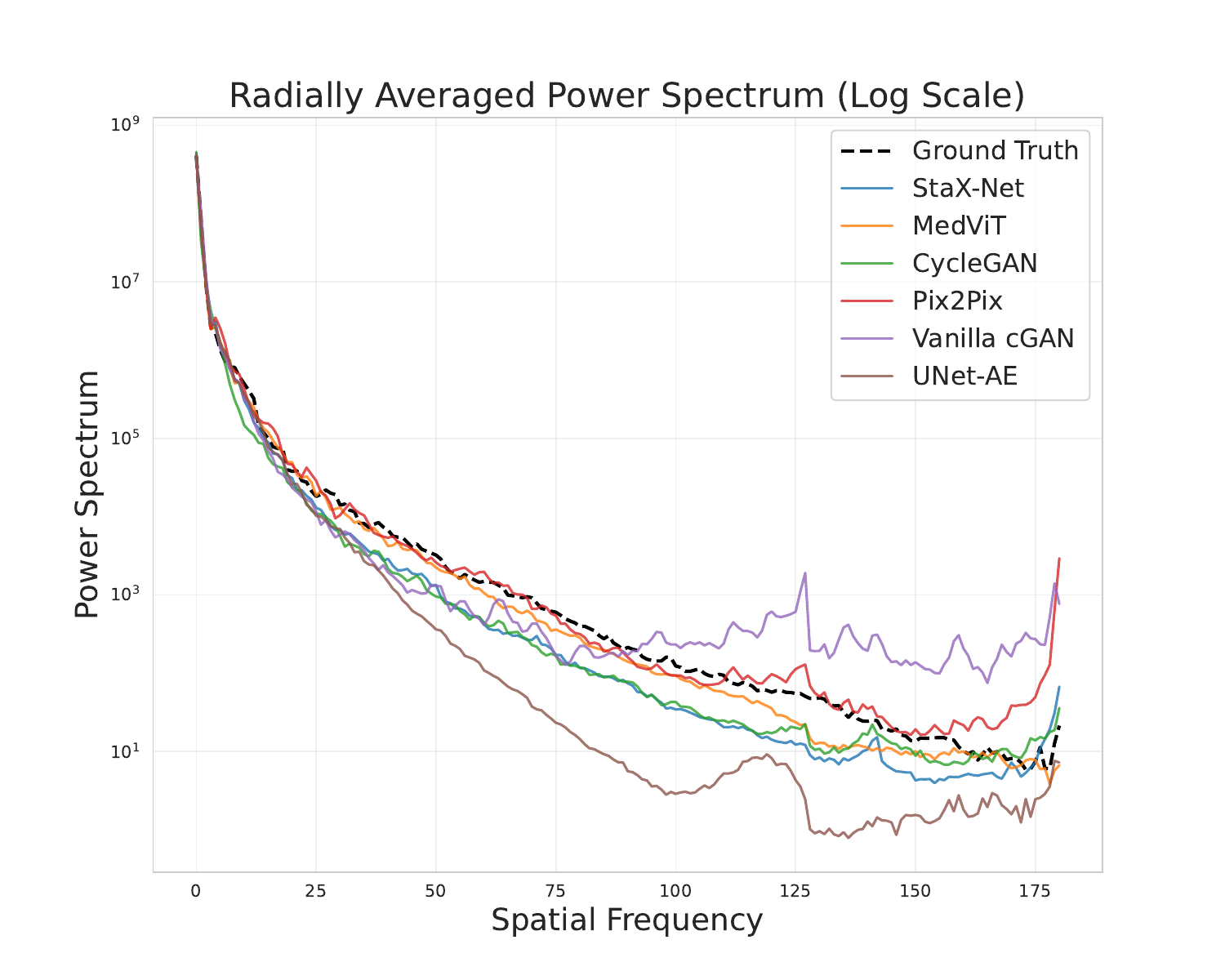} \\
        
        (a) & (b) & (c) \\
        
    \end{tabular}
    \caption{1D radially averaged Fourier power spectrum analysis for (a) Head \& Neck, (b) Thorax, and (c) Abdomen/Pelvis datasets. The x-axis represents spatial frequency, while the y-axis shows the logarithmic power magnitude.}
    \label{fig:fft_spectrum}
\end{figure*}

The deterministic UNet-AE shows reduced high spatial-frequency content relative to the GT, consistent with its smoother reconstructions. Vanilla cGAN exhibits larger deviations in selected high-frequency regions. StaX-Net follows the GT trajectory more closely across the evaluated spatial-frequency range, indicating preservation of frequency characteristics relevant to fine structural texture, as demonstrated in Fig. \ref{fig:fft_spectrum}.

\subsubsection{Volumetric (3D) Stress Test and SOTA Comparison}

A pervasive claim in recent literature is that 2D generative models inherently fail to preserve z-axis continuity, motivating computationally expensive 3D architectures. To investigate this, we conducted a volumetric stress test measuring the z-axis gradient mean absolute error (Z-MAE) and 3D-SSIM.

StaX-Net demonstrates competitive cross-slice consistency in the evaluated volumetric metrics. Because its source representation is processed through the shared blocks and FAST prior, the sequential 2D outputs can preserve volumetric characteristics without explicit 3D convolutions. StaX-Net is also compared with the 3D diffusion framework MD-DGA, which reported a peak PSNR of $16.64$ dB on comparable data. These results indicate that the proposed 2D framework can provide competitive volumetric fidelity while avoiding the computational and memory requirements of explicit 3D diffusion processing.

\section{Conclusion}
Here, we have presented StaX-Net, a Stackelberg-inspired interleaved asymmetric framework for CT-to-MRI translation. The interleaved shared blocks between $G$ and $D$ enhance feature alignment, while explicit parameter freezing and asymmetric learning rates stabilize the alternating leader-follower updates. Furthermore, the LUPI-based FAST layer acts as a feature transformer that distills target-supervised spatial priors into a dynamic, source-only mapping for inference. 

Through comprehensive multi-dimensional evaluations we have highlighted the inadequacy of relying solely on classical pixel-wise metrics for medical image translation. For instance, while deterministic baselines like UNet-AE achieved competitive PSNR scores, frequency-domain and deep perceptual analyses revealed their tendency to produce over-smoothed approximations lacking essential clinical texture. In contrast, StaX-Net consistently demonstrated a superior balance, effectively preserving both global anatomical geometry and fine-grained, high-frequency structural details. By constraining the synthesis process through these target-informed representations, StaX-Net mitigates the risk of anatomically implausible hallucinations relative to standard adversarial models, thereby offering a more robust and safe foundation for reliable clinical diagnosis.

\section*{Data Availability Statement}
\sloppy
The original SynthRAD 2025 Grand Challenge dataset is officially hosted at \underline{https://synthrad2025.grand-challenge.org/}. The specific data partition and format utilized in this study can be openly accessed via Kaggle at \underline{https://www.kaggle.com/datasets/hdhduxhd/synthrad2025-task1}.

\section*{Declaration of Generative AI in Scientific Writing}
During the preparation of this work, the authors used Gemini Pro solely to improve the English language, correct grammar, and enhance the overall flow of the text. After using this tool, the authors reviewed and edited the content as needed and take full responsibility for the final content of the publication.

\bibliographystyle{elsarticle-num}
\bibliography{references}

\end{document}